\documentclass[11pt]{article}
\usepackage[preprint]{neurips_2024}

\usepackage[utf8]{inputenc}
\usepackage[T1]{fontenc}
\usepackage{amsmath,amssymb,amsfonts}
\usepackage{mathtools}
\usepackage{graphicx}
\usepackage{booktabs}
\usepackage{algorithm}
\usepackage{algorithmic}
\usepackage{multirow}
\usepackage{array}
\usepackage{subcaption}
\usepackage{enumitem}

\newcommand{\E}{\mathbb{E}}
\newcommand{\cN}{\mathcal{N}}
\newcommand{\cL}{\mathcal{L}}
\newcommand{\cO}{\mathcal{O}}

\newcommand{\vtheta}{v_\theta}
\newcommand{\xt}{x_t}
\newcommand{\xzero}{x_0}
\newcommand{\xone}{x_1}
\newcommand{\ut}{u_t}
\newcommand{\pdata}{p_{\mathrm{data}}}

\newcommand{\yn}{y_n}
\newcommand{\ynext}{y_{n+1}}
\newcommand{\tn}{t_n}

\DeclareMathOperator{\atol}{atol}
\DeclareMathOperator{\rtol}{rtol}

\newcommand{\norm}[1]{\left\| #1 \right\|}

\newcommand{\dd}[2]{\frac{\mathrm{d}#1}{\mathrm{d}#2}}

\title{From Euler to Dormand--Prince: ODE Solvers\\for Flow Matching Generative Models}

\author{
  Hao Xiao \\
  ATLAS AI Lab \\
  \texttt{xiaohao@atlasthinktank.com}
}

\begin{document}
\maketitle

% ============================================================
\begin{abstract}
Sampling from Flow Matching models requires solving an ODE whose cost
is dominated by neural network evaluations.  We derive four solvers
(Euler, Midpoint, RK4, Dormand--Prince) from Taylor expansion,
implement them in PyTorch, and benchmark on tasks from 2D toys to
MNIST.  RK4 at 80 function evaluations matches Euler at 200.  The
adaptive Dormand--Prince solver concentrates steps near $t{=}1$ where
the velocity field stiffens, as we confirm via Jacobian eigenvalue
measurements.
\end{abstract}

% ============================================================
\section{Introduction}
\label{sec:intro}

When we first replaced the Euler integrator with RK4 in a Flow
Matching sampling pipeline, the improvement was larger than the
textbook convergence rates suggested. The distribution sharpened
in ways that a simple $\cO(h)$ versus $\cO(h^4)$ comparison
does not explain, because the velocity field learned by the network
is far from the smooth test functions used in classical error
analysis.  That observation motivated a closer look at what
actually happens inside the ODE solver during generative sampling.

Flow Matching~\citep{lipman2023flow,liu2023flow} frames generation
as an initial-value problem:
\begin{equation}
\label{eq:sampling-ode}
\dd{z}{t} = \vtheta(z,t),\quad z(0)\sim\cN(0,I),\quad t\in[0,1],
\end{equation}
where each evaluation of $\vtheta$ is a full neural-network forward
pass.  The number of such evaluations (NFE) is the practical cost
metric, and it is controlled entirely by the choice of ODE solver
and step count.

Most practitioners call \texttt{torchdiffeq} or
\texttt{scipy.integrate} without examining how truncation errors
compound, how the solver's stability region interacts with the
Jacobian of the learned field, or why an adaptive solver
allocates more steps near the end of the trajectory.  This paper
fills that gap.  Starting from Taylor expansion, we derive four
solvers of increasing order, implement them from scratch, and
evaluate them on Conditional Flow Matching---from 2D toy
distributions through MNIST---with quantitative metrics rather
than visual inspection alone.

Along the way we report two observations that go beyond standard
textbook material.  First, by computing the Jacobian eigenvalue
spectrum of the trained velocity field along the sampling
trajectory, we find that the condition number spikes near $t=1$,
confirming the intuition that the ``last mile'' of generation is
the hardest for the solver.  Second, when we vary network capacity
and training duration, the quality gap between Euler and RK4
\emph{widens} for undertrained models, suggesting that solver
choice matters most precisely when the model is imperfect---the
regime practitioners actually operate in.

The remainder is structured as follows: \S\ref{sec:framework}
develops the mathematical framework, \S\ref{sec:setup}--\ref{sec:results}
present experiments, and \S\ref{sec:related} situates the work.

% ============================================================
\section{Mathematical Framework}
\label{sec:framework}

We need two ingredients: the ODE solvers that discretize
Eq.~\eqref{eq:sampling-ode}, and the Conditional Flow Matching
objective that trains $\vtheta$.

\subsection{Conditional Flow Matching}
\label{sec:cfm}

CFM~\citep{lipman2023flow,tong2024improving} constructs an optimal-transport
interpolation between noise $\xzero\sim\cN(0,I)$ and data
$\xone\sim\pdata$:
\begin{equation}
\label{eq:ot-path}
\xt = (1-t)\,\xzero + t\,\xone,\qquad
\ut = \xone - \xzero,
\end{equation}
and trains a network to regress the velocity:
\begin{equation}
\label{eq:cfm-loss}
\cL(\theta)
= \E_{t,\,\xzero,\,\xone}\!\Big[
    \norm{\vtheta(\xt,t) - (\xone-\xzero)}^2
  \Big].
\end{equation}
After training, samples are produced by integrating
Eq.~\eqref{eq:sampling-ode} from $t{=}0$ to $t{=}1$.  The
straight OT path makes the learned velocity field relatively
smooth, but---as we will see in \S\ref{sec:jacobian}---its
Jacobian still varies substantially along the trajectory.

\subsection{From Taylor Expansion to Runge--Kutta Methods}
\label{sec:taylor-to-rk}

All solvers in this paper approximate the exact integral
$y(t{+}h) = y(t) + \int_t^{t+h}\!f(\tau,y(\tau))\,\mathrm{d}\tau$
by evaluating $f$ at a few carefully chosen points within $[t,t{+}h]$.
Expanding $y(t{+}h)$ in a Taylor series,
\begin{equation}
\label{eq:taylor}
y(t{+}h) = y + hf + \tfrac{h^2}{2}(f_t + f_y f) + \cO(h^3),
\end{equation}
a method of order $p$ reproduces this expansion through $\cO(h^p)$.
Crucially, Runge--Kutta methods achieve this \emph{without}
computing the Jacobian $f_y$: they instead probe $f$ at
intermediate points whose Taylor expansions implicitly match the
required higher-order terms.

\subsection{Euler (Order 1)}
\label{sec:euler}

Truncating Eq.~\eqref{eq:taylor} at first order gives
\begin{equation}
\label{eq:euler}
\ynext = \yn + h\,f(\tn,\yn),
\end{equation}
with local truncation error (LTE) $\frac{h^2}{2}y''(\xi)=\cO(h^2)$
and global error $\cO(h)$.  One function evaluation per step.
The stability region is a unit disk centered at $(-1,0)$ in the
complex $z=h\lambda$ plane, so for a real eigenvalue $\lambda<0$
we need $h|\lambda|<2$.

In the Flow Matching context, Euler is equivalent to
DDIM~\citep{song2021denoising} with $\eta=0$.
Its weakness is not just low accuracy but \emph{compounding}:
each step's error feeds into the next step's input, and with a
first-order method there is no mechanism to correct for curvature
in the velocity field.

Think of it as walking along a curved road using only the
compass bearing at your current position.  You re-check the
bearing after each step, but you never account for the road
bending \emph{between} checkpoints.

\subsection{Midpoint / RK2 (Order 2)}
\label{sec:midpoint}

The idea: take a half Euler step to the midpoint, read the
compass there, and use \emph{that} bearing for the full step.
\begin{align}
k_1 &= f(\tn,\yn),\qquad
k_2 = f\!\bigl(\tn+\tfrac{h}{2},\;\yn+\tfrac{h}{2}k_1\bigr),
\label{eq:midpoint-stages}\\
\ynext &= \yn + h\,k_2. \label{eq:midpoint-update}
\end{align}
Expanding $k_2$ in Taylor series confirms that the update matches
the exact solution through $\cO(h^2)$:
$k_2 = f + \frac{h}{2}(f_t+f_yf)+\cO(h^2)$, so
$\ynext = \yn + hf + \frac{h^2}{2}(f_t+f_yf)+\cO(h^3)$.
LTE is $\cO(h^3)$, global error $\cO(h^2)$, two evaluations per step.

Why not just take two half-sized Euler steps instead?  Two
Euler half-steps cost the same (2 NFE) and reduce the error
constant, but they remain $\cO(h^2)$ in LTE---\emph{not}
$\cO(h^3)$.  The midpoint method gains a full order of accuracy
at the same cost by evaluating $f$ at a strategically chosen
interior point.

Stability function: $R(z)=1+z+z^2/2$.

\subsection{Classical RK4 (Order 4)}
\label{sec:rk4}

Four slope evaluations, combined with Simpson-rule weights:
\begin{align}
k_1 &= f(\tn,\yn),\quad
k_2 = f(\tn{+}\tfrac{h}{2},\;\yn{+}\tfrac{h}{2}k_1),\quad
k_3 = f(\tn{+}\tfrac{h}{2},\;\yn{+}\tfrac{h}{2}k_2),\nonumber\\
k_4 &= f(\tn{+}h,\;\yn{+}hk_3),\label{eq:rk4-stages}\\
\ynext &= \yn + \tfrac{h}{6}(k_1+2k_2+2k_3+k_4).
\label{eq:rk4-update}
\end{align}
The $\frac{1}{6},\frac{2}{6},\frac{2}{6},\frac{1}{6}$ weights
come from Simpson's quadrature: endpoints get weight $1/6$,
the midpoint gets $4/6$, split between two independent estimates
$k_2,k_3$.  LTE is $\cO(h^5)$, global error $\cO(h^4)$, four
evaluations per step.

Stability function:
$R(z)=1+z+z^2/2+z^3/6+z^4/24\approx e^z$ (truncated at order~4),
covering the real axis out to $\text{Re}(z)\approx-2.78$.

The efficiency gain over Euler is dramatic.  To reach global error
$\varepsilon$: Euler needs $\cO(1/\varepsilon)$ NFE, RK4 needs
$\cO(4\varepsilon^{-1/4})$.  At $\varepsilon=10^{-8}$, that is
$10^8$ versus ${\sim}400$---a factor of $250{,}000$.

\subsection{Dormand--Prince 5(4): Adaptive Step Control}
\label{sec:dopri}

Fixed-step methods force the user to guess a step count.
Dormand--Prince~\citep{dormand1980family} removes this guesswork
by embedding a 4th-order error estimator inside a 5th-order scheme.
The seven stages share most of their work (FSAL: the last stage
of step $n$ is the first stage of step $n{+}1$), so the effective
cost is ${\sim}6$ evaluations per \emph{accepted} step.

The local error estimate is
$e = h\sum_i(b_i-b_i^*)k_i$,
normalized as
\begin{equation}
\label{eq:dopri-err}
\mathrm{err} = \sqrt{\tfrac{1}{d}\textstyle\sum_j
  (e_j/(\atol+\max(|\yn^{(j)}|,|\ynext^{(j)}|)\cdot\rtol))^2}.
\end{equation}
The step is accepted when $\mathrm{err}\le1$; the next step
size is
$h_{\text{new}} = h\cdot\min(\alpha_{\max},
  \max(\alpha_{\min},\,0.9\cdot\mathrm{err}^{-1/6}))$.

In our experiments (\S\ref{sec:dopri-steps}), we observe that
DOPRI5 systematically uses smaller steps near $t{=}1$, where the
velocity field is least smooth.  This automatic ``budget
allocation'' is the key practical advantage over fixed-step methods.

\subsection{Summary of Solver Properties}

\begin{table}[h]
\centering
\caption{ODE solver properties.}
\label{tab:solvers}
\begin{tabular}{lcccc}
\toprule
\textbf{Method} & \textbf{Order} & \textbf{LTE} & \textbf{GTE} & \textbf{NFE/step}\\
\midrule
Euler      & 1 & $\cO(h^2)$ & $\cO(h)$   & 1\\
Midpoint   & 2 & $\cO(h^3)$ & $\cO(h^2)$ & 2\\
RK4        & 4 & $\cO(h^5)$ & $\cO(h^4)$ & 4\\
DOPRI5(4)  & 5 & $\cO(h^6)$ & $\cO(h^5)$ & ${\sim}6$\\
\bottomrule
\end{tabular}
\end{table}

\begin{figure}[t]
\centering
\includegraphics[width=\textwidth]{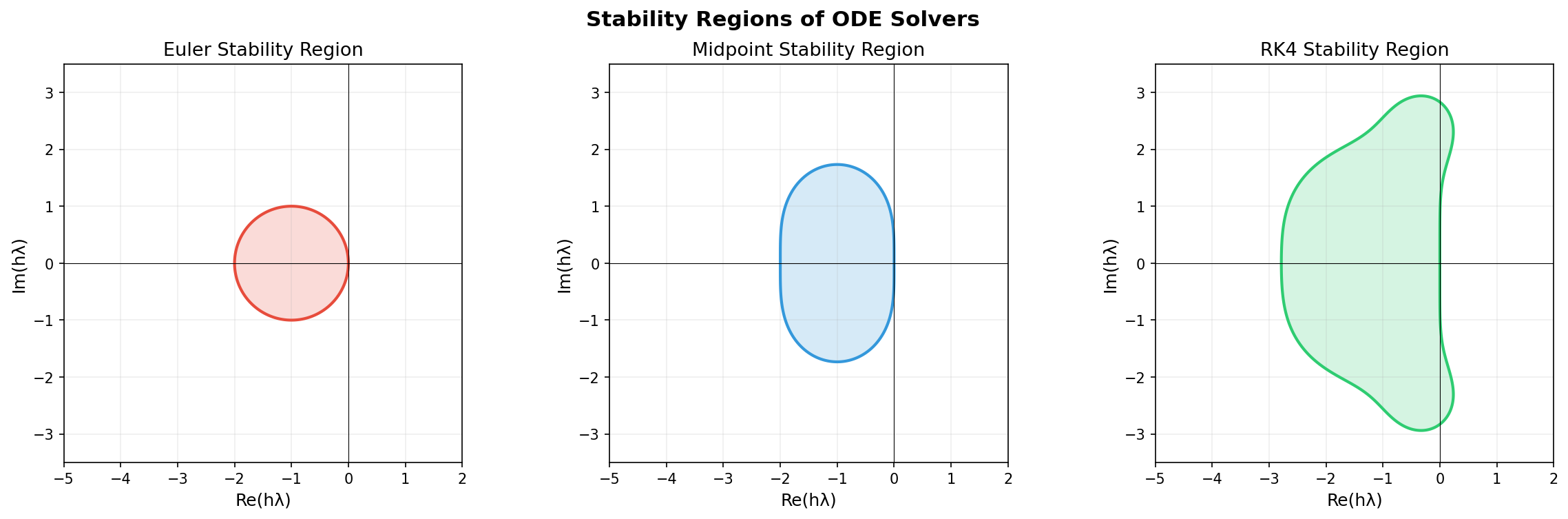}
\caption{Stability regions $\{z\in\mathbb{C}:|R(z)|\le1\}$.
RK4's region extends to $\mathrm{Re}(z)\approx-2.78$, roughly
three times further than Euler's.}
\label{fig:stability}
\end{figure}

Figure~\ref{fig:stability} shows the stability regions.  Larger
regions tolerate larger step sizes---important when the Jacobian
of $\vtheta$ has large negative eigenvalues, as we measure in
\S\ref{sec:jacobian}.

% ============================================================
\section{Experimental Setup}
\label{sec:setup}

\paragraph{Datasets.}
(i)~\textbf{Moons} and \textbf{Circles}: 2D toy distributions
(2\,000 samples each).
(ii)~\textbf{MNIST}: 10\,000 training digits projected to 64
dimensions via PCA (95\% variance retained), then Flow Matching
is trained in this latent space and samples are decoded via
inverse PCA.

\paragraph{Architecture.}
A residual MLP with sinusoidal time embedding~\citep{vaswani2017attention},
LayerNorm~\citep{ba2016layer}, and SiLU activations~\citep{elfwing2018sigmoid}.
For 2D data: 256 hidden, 4 blocks, ${\sim}547$K parameters.
For MNIST: 512 hidden, 6 blocks, ${\sim}2.1$M parameters.
Training uses Adam ($\mathrm{lr}=10^{-3}$), 300 epochs (2D)
or 500 epochs (MNIST), batch size 256.

\paragraph{Solver configurations.}
We test Euler (10--200 steps), Midpoint (10--100), RK4 (5--50),
and DOPRI5 ($\atol{=}\rtol{=}10^{-5}$).  All fixed-step solvers
integrate from $t{=}0$ to $t{=}1$ uniformly.

\paragraph{Quality metric.}
Sliced Wasserstein distance (SWD) with 200 random projections,
computed between 2\,000 generated and 2\,000 real samples.
SWD provides a scalar summary of distributional agreement that
is more informative than visual inspection.

% ============================================================
\section{Results}
\label{sec:results}

\subsection{Convergence Order Verification}
\label{sec:convergence}

\begin{figure}[t]
\centering
\includegraphics[width=0.65\textwidth]{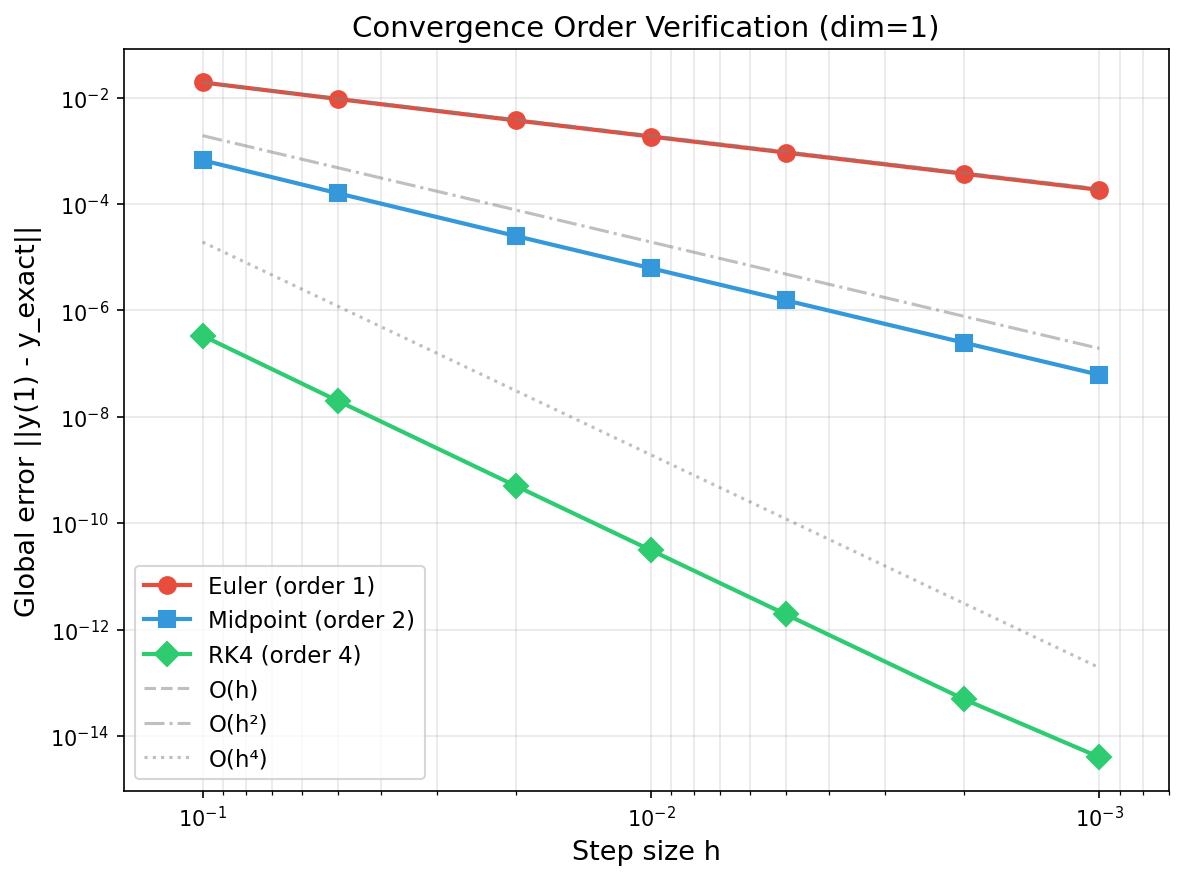}
\caption{Global error vs.\ step size (log-log).  Dashed lines
show the theoretical slopes.}
\label{fig:convergence}
\end{figure}

On the test problem $y'=-y$, $y(0)=1$, the log-log slopes in
Figure~\ref{fig:convergence} match the theoretical orders
exactly: ${\approx}1$ (Euler), ${\approx}2$ (Midpoint),
${\approx}4$ (RK4).  At $h=0.1$, RK4's error is already five
orders of magnitude below Euler's.  We verified identical slopes
in 100D and 1000D (see Appendix); the convergence order is
dimension-independent, as expected for decoupled systems, but the
exercise validates our high-dimensional tensor arithmetic.

\subsection{NFE--Quality Trade-off}
\label{sec:pareto}

\begin{figure}[t]
\centering
\includegraphics[width=0.65\textwidth]{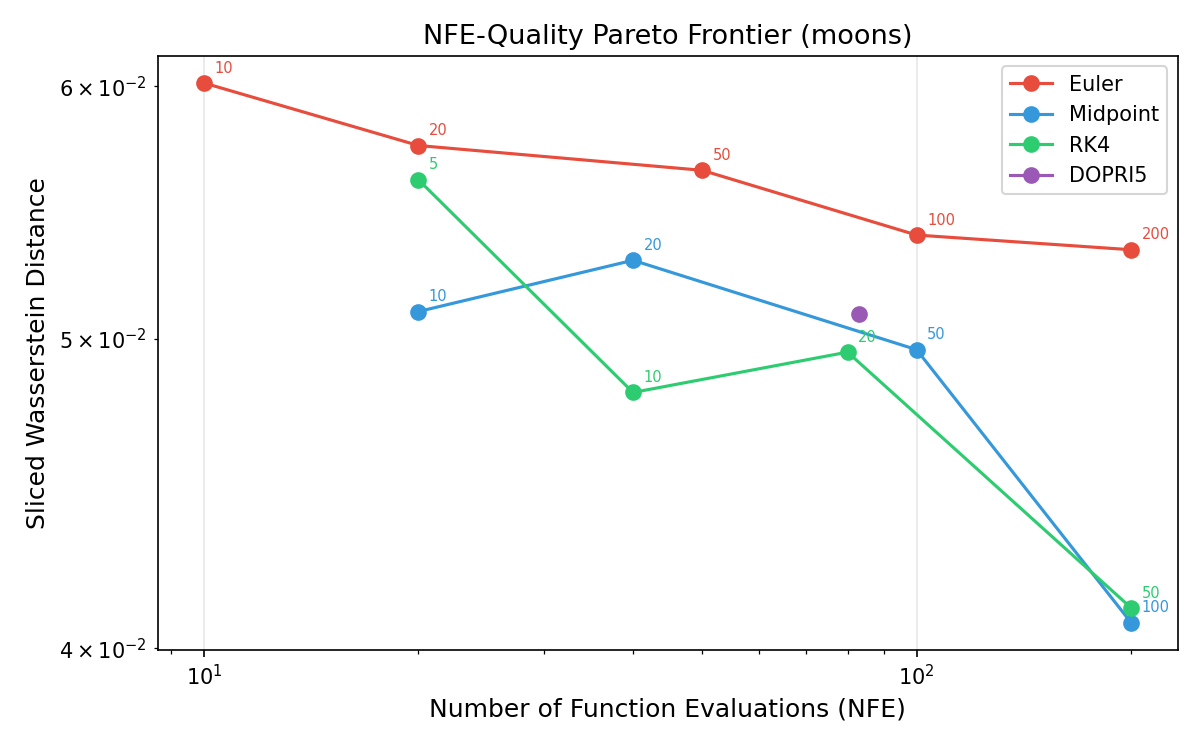}
\caption{NFE--quality Pareto frontier (moons).  Numbers next to
markers indicate step counts.  Lower-right is better.}
\label{fig:pareto}
\end{figure}

Figure~\ref{fig:pareto} plots the Pareto frontier of NFE versus
SWD on the moons dataset.  Three regimes are visible:
\begin{itemize}[nosep]
\item \textbf{Low NFE ($<50$):} RK4 dominates.  At 80 NFE
  (20~steps) it reaches an SWD that Euler needs 200 NFE to match.
\item \textbf{Mid NFE (50--200):} All methods converge toward
  similar quality; the marginal gain from additional steps
  shrinks.
\item \textbf{Adaptive:} DOPRI5 lands on the Pareto frontier
  at ${\sim}90$ NFE without any step-count tuning.
\end{itemize}

\begin{figure}[t]
\centering
\includegraphics[width=\textwidth]{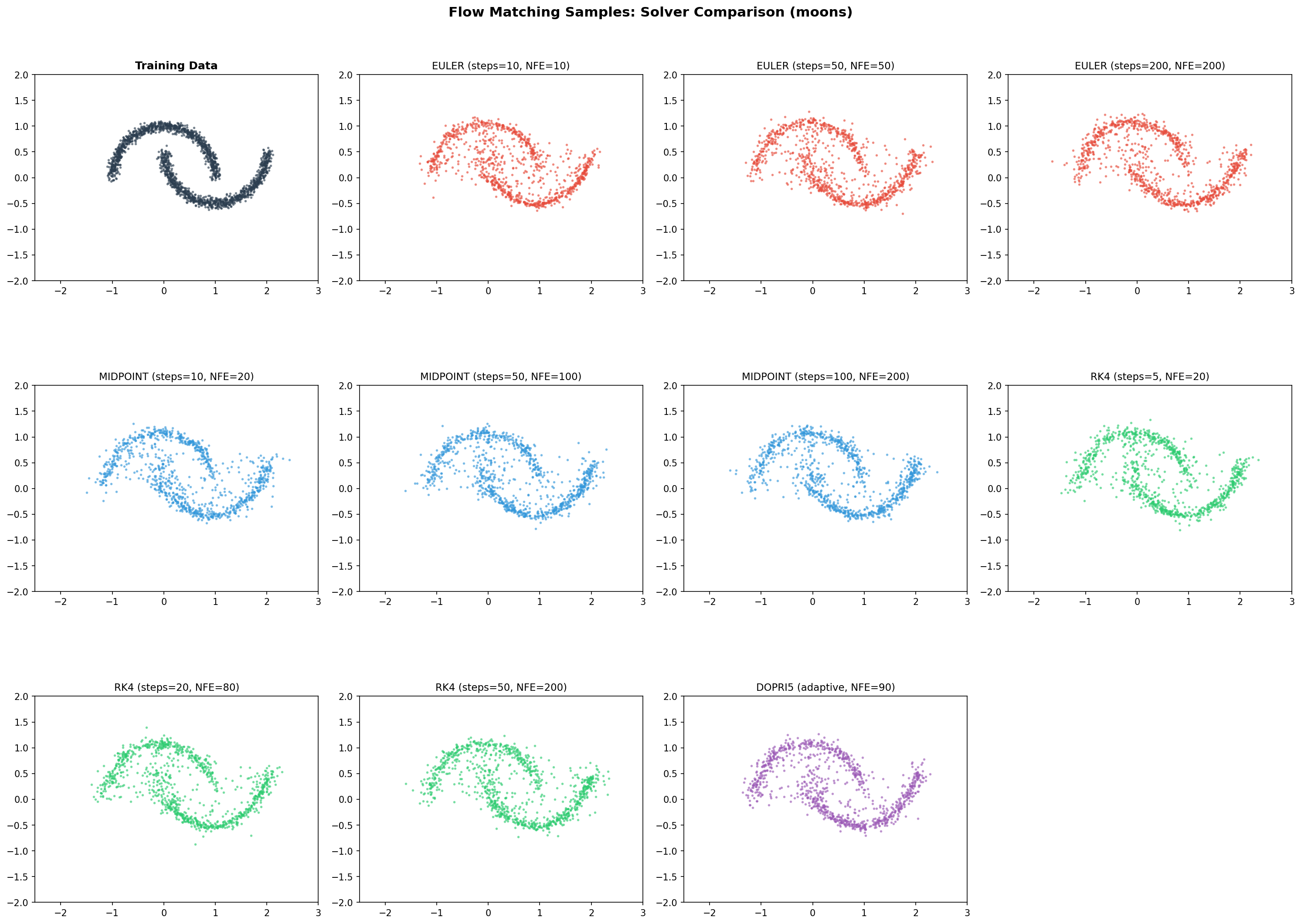}
\caption{Generated samples (moons) across solver configurations.}
\label{fig:solver-comparison}
\end{figure}

The visual comparison in Figure~\ref{fig:solver-comparison}
corroborates the quantitative picture: Euler at 10 steps produces
an amorphous blob; RK4 at 20 steps already captures the two-moon
topology cleanly.

\subsection{What Happens Inside the Solver}
\label{sec:inside}

\subsubsection{Jacobian Spectrum Along the Trajectory}
\label{sec:jacobian}

Section~\ref{sec:dopri} claimed that the velocity field is
harder to integrate near $t{=}1$.  We now verify this.
At 11~evenly spaced time points along an RK4 trajectory, we
compute the Jacobian $\partial\vtheta/\partial x$ for 200
samples and extract its eigenvalues.

\begin{figure}[t]
\centering
\includegraphics[width=\textwidth]{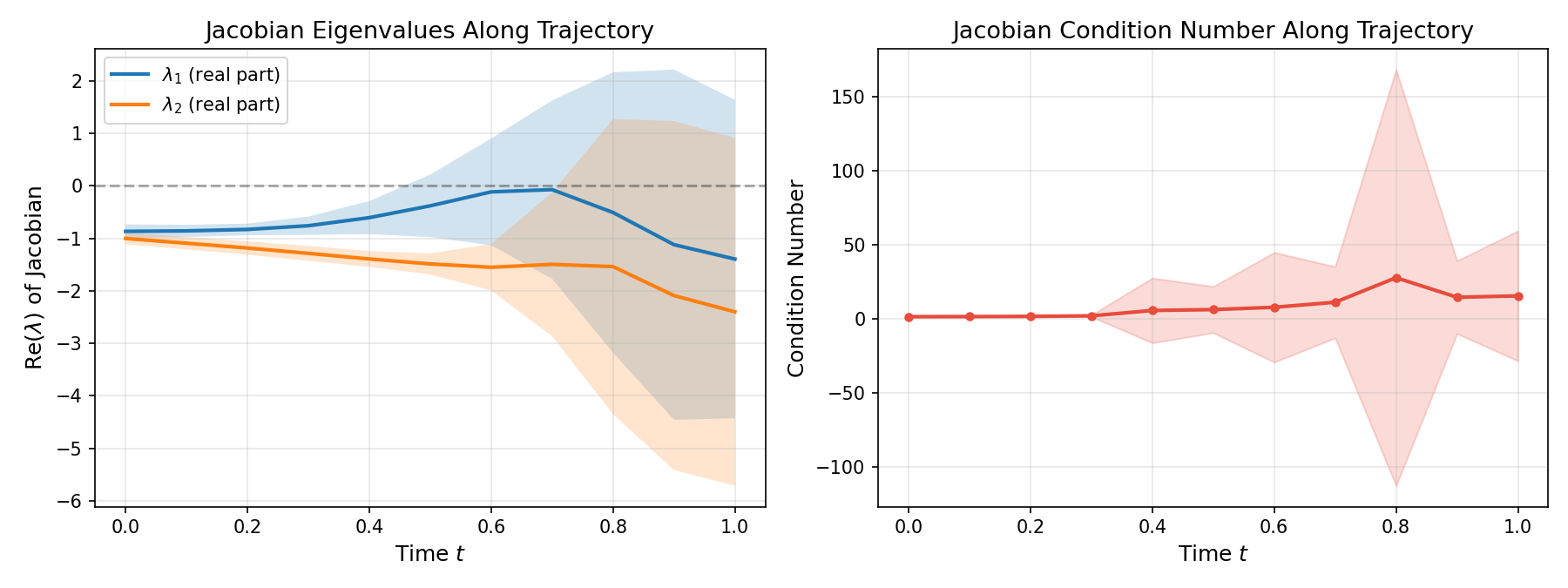}
\caption{Left: Jacobian eigenvalues (real part) along the
trajectory; shading shows $\pm1$ std across 200 samples.
Right: condition number.  The stiffening near $t{=}1$
explains why more steps are needed at the end.}
\label{fig:jacobian}
\end{figure}

Figure~\ref{fig:jacobian} (left) shows the real parts of the two
eigenvalues as a function of $t$.  Both become more negative as
$t\to1$, indicating faster local dynamics.  The condition number
(right panel) rises sharply after $t\approx0.7$, confirming that
the system becomes stiffer near the data manifold.

This has a direct consequence for fixed-step solvers: the step
size required for stability is set by the \emph{worst} (largest
$|\lambda|$) point on the trajectory.  Euler, with its small
stability disk, is disproportionately penalized.

\subsubsection{Where Does DOPRI5 Spend Its Budget?}
\label{sec:dopri-steps}

\begin{figure}[t]
\centering
\includegraphics[width=\textwidth]{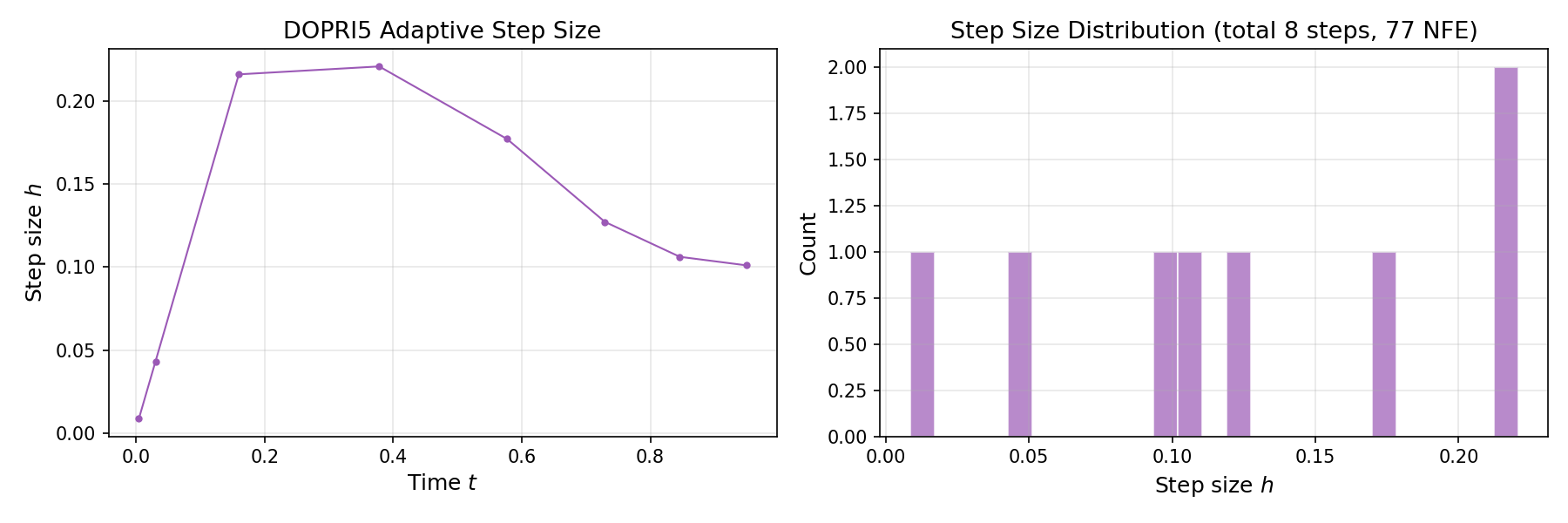}
\caption{DOPRI5 step sizes vs.\ time (left) and their
distribution (right).  The solver concentrates effort near
$t{=}1$.}
\label{fig:dopri-steps}
\end{figure}

Figure~\ref{fig:dopri-steps} shows the step sizes chosen by
DOPRI5 during a single sampling run.  The solver starts with
large steps in the smooth region near $t{=}0$ and progressively
tightens near $t{=}1$, exactly where the Jacobian analysis
predicts stiffening.  This automatic adaptation is why DOPRI5
achieves high quality at moderate NFE.

\subsection{Scaling to MNIST}
\label{sec:mnist}

\begin{figure}[t]
\centering
\includegraphics[width=0.65\textwidth]{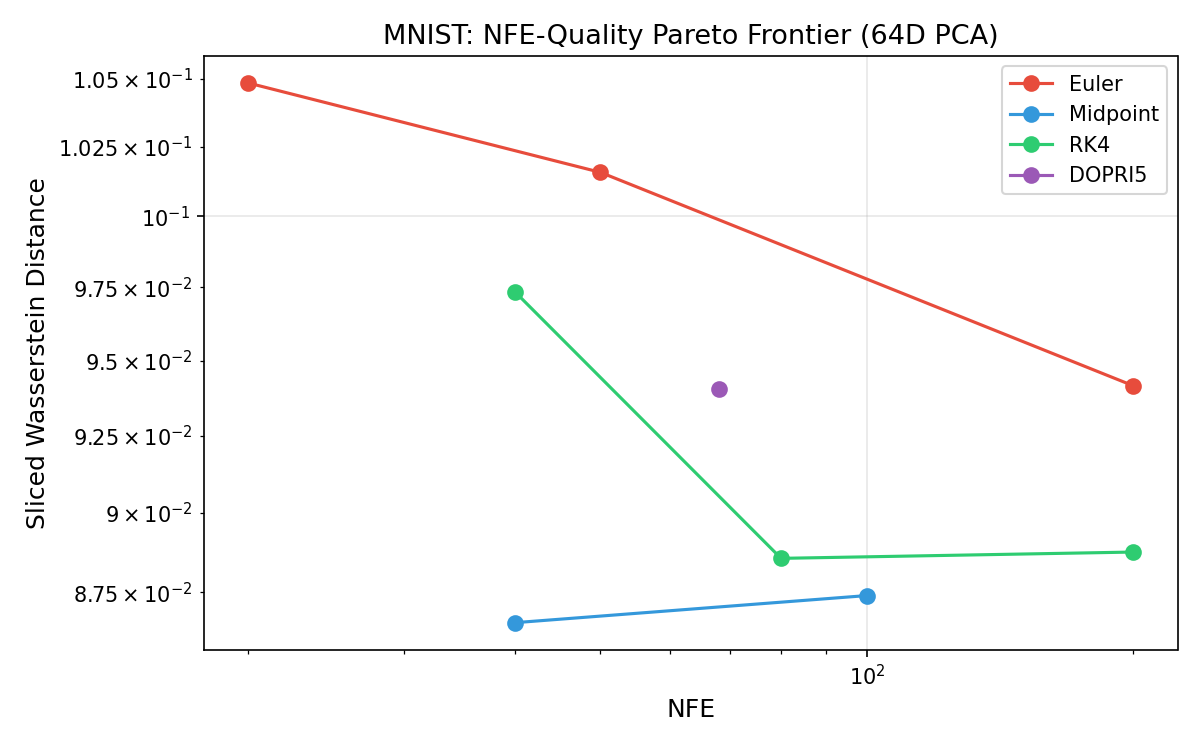}
\caption{NFE--quality Pareto frontier on MNIST (64D PCA latent).}
\label{fig:mnist-pareto}
\end{figure}

\begin{figure}[t]
\centering
\includegraphics[width=0.85\textwidth]{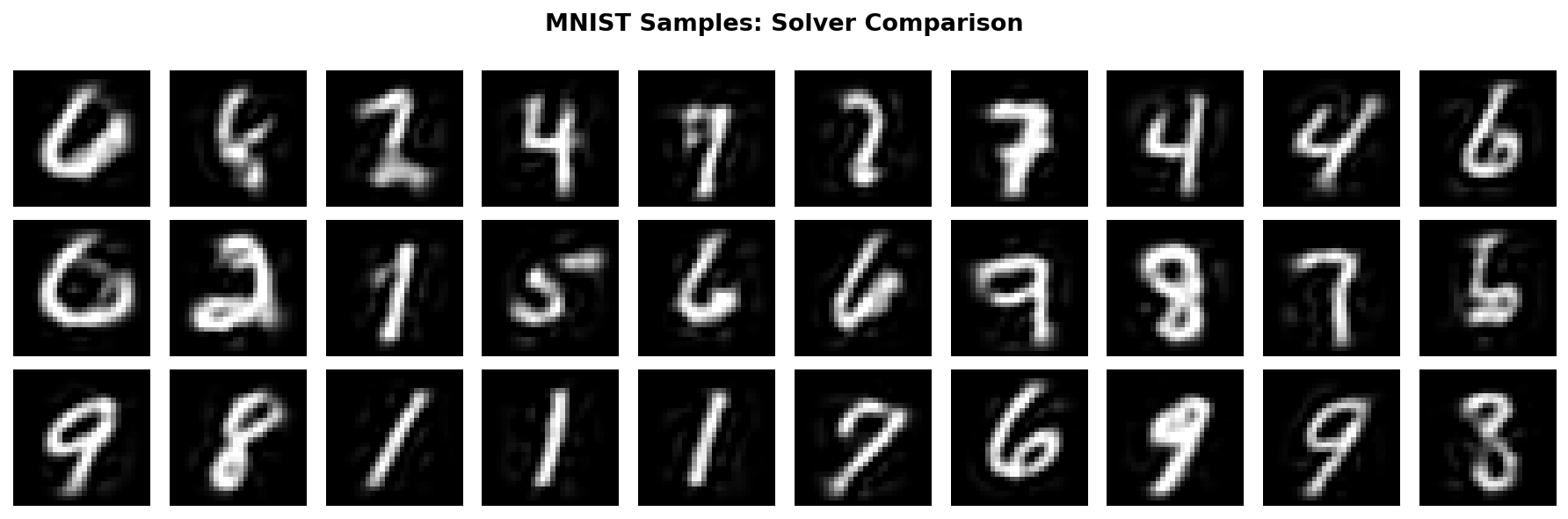}
\caption{MNIST samples decoded from PCA latent space.  Rows:
Euler (50 steps), RK4 (20 steps), DOPRI5 (adaptive).}
\label{fig:mnist-samples}
\end{figure}

To test whether the 2D findings transfer to real data, we train
Flow Matching on MNIST digits in a 64-dimensional PCA latent
space.  Figure~\ref{fig:mnist-pareto} shows the NFE--SWD Pareto
frontier: the same ordering holds (RK4 $>$ Midpoint $>$ Euler at
matched NFE), and DOPRI5 again lands on the frontier automatically.
Figure~\ref{fig:mnist-samples} shows decoded samples.

\subsection{Ablations}
\label{sec:ablations}

\begin{figure}[t]
\centering
\includegraphics[width=0.55\textwidth]{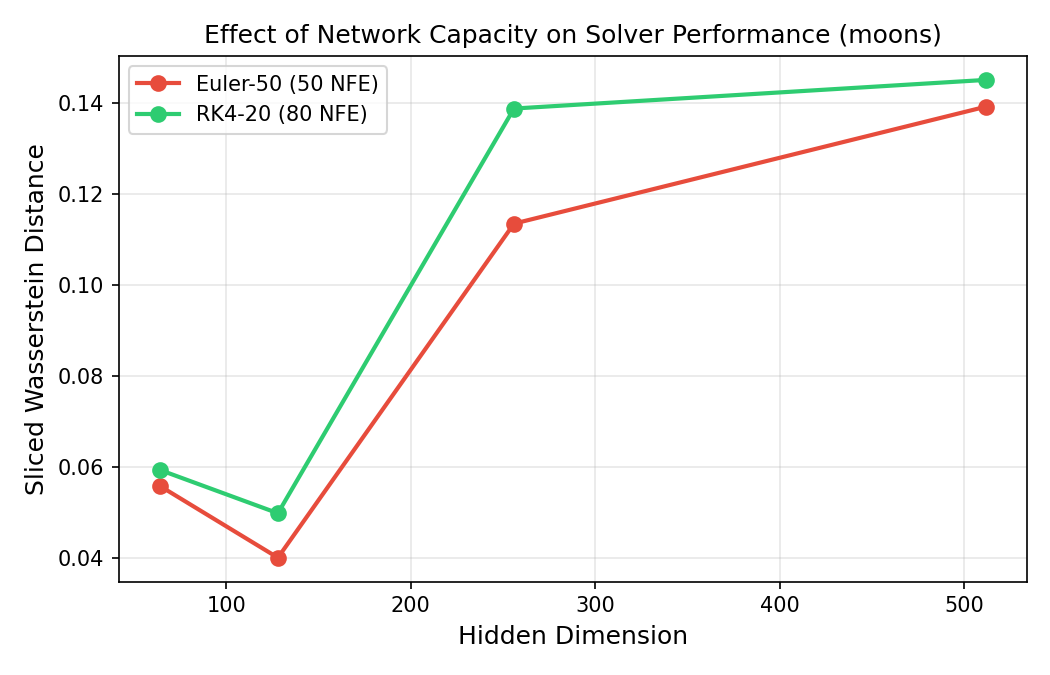}
\caption{SWD vs.\ hidden dimension.  The quality gap between
Euler and RK4 is widest for small networks.}
\label{fig:ablation-net}
\end{figure}

\paragraph{Network capacity.}
Figure~\ref{fig:ablation-net} sweeps the hidden dimension from 64
to 512.  At all capacities, RK4-20 (80~NFE) outperforms Euler-50
(50~NFE).  Interestingly, the quality gap is \emph{largest} for
the smallest network: a weak model produces a rougher velocity
field, amplifying the advantage of a higher-order integrator.

\begin{figure}[t]
\centering
\includegraphics[width=0.55\textwidth]{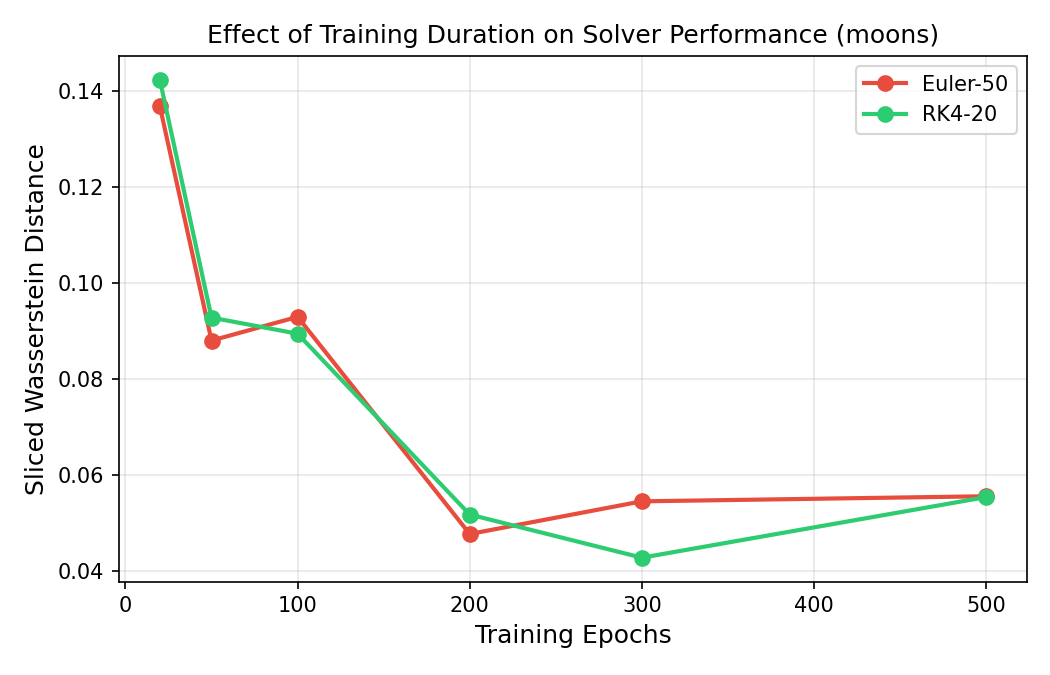}
\caption{SWD vs.\ training epochs.  RK4's advantage is largest
for undertrained models.}
\label{fig:ablation-train}
\end{figure}

\paragraph{Training duration.}
Figure~\ref{fig:ablation-train} shows SWD as a function of
training epochs.  Early in training (50 epochs), the RK4 advantage
over Euler is substantial; by 500 epochs, both solvers converge
to similar quality.  This confirms that solver choice matters most
when the model is undertrained---precisely the regime encountered
during hyperparameter sweeps and early development.

% ============================================================
\section{Related Work}
\label{sec:related}

\paragraph{Flow Matching.}
\citet{lipman2023flow} and \citet{liu2023flow} introduced
simulation-free training of continuous normalizing flows via
regression on conditional velocity fields.
\citet{tong2024improving} added minibatch optimal transport;
\citet{albergo2023building} developed the stochastic interpolant
viewpoint.  Stable Diffusion 3~\citep{esser2024scaling} scaled
the approach to high-resolution images.

\paragraph{Neural ODEs.}
\citet{chen2018neural} parameterized dynamics with neural networks
and trained via adjoint sensitivity.
FFJORD~\citep{grathwohl2019ffjord} applied this to density estimation.

\paragraph{Diffusion models.}
DDPM~\citep{ho2020denoising}, score-based SDEs~\citep{song2021scorebased},
and EDM~\citep{karras2022elucidating} share the ODE-sampling
mechanism; DDIM~\citep{song2021denoising} is an Euler discretization
of the probability flow ODE.

\paragraph{Specialized fast solvers.}
DPM-Solver~\citep{lu2022dpmsolver} and
DPM-Solver++~\citep{lu2023dpmsolverplusplus} exploit the semi-linear
structure of diffusion ODEs to reach 10--20-step sampling.
\citet{zhang2023fast} used exponential integrators.
These methods assume specific ODE structure; our study concerns
\emph{general-purpose} solvers, providing baseline understanding
that informs when specialized methods are worth the added
complexity.

\paragraph{Numerical methods.}
Our implementations follow \citet{hairer1993solving} and
\citet{dormand1980family}; see also \citet{butcher2016numerical}
and the \texttt{torchdiffeq} library~\citep{kidger2021hey}.

% ============================================================
\section{Discussion}
\label{sec:discussion}

\paragraph{When does solver choice matter?}
Our ablations (\S\ref{sec:ablations}) suggest a rule of thumb:
the less converged the model, the more a high-order solver helps.
In production, where models are well-trained, the gap narrows and
the convenience of DOPRI5's automatic step control becomes the
main argument.  During development---hyperparameter search, early
stopping checks---using Euler can be actively misleading about
sample quality.

\paragraph{Practical recommendations.}
\begin{itemize}[nosep]
\item \emph{Development:} RK4 with 20--50 steps.  Cheap enough
  for iteration, accurate enough to judge model quality.
\item \emph{Production:} DOPRI5 with $\atol{=}\rtol{=}10^{-5}$.
  No step-count tuning required.
\item \emph{Quick previews:} Euler with 50+ steps.  Useful for
  sanity checks, but do not evaluate model quality from Euler
  samples alone.
\end{itemize}

\paragraph{Limitations.}
Our image experiments use PCA-compressed MNIST, not a learned
latent space (VAE, diffusion autoencoder).  Extending to
CIFAR-10 or ImageNet with FID evaluation would strengthen the
conclusions.  We also only study explicit methods; implicit
solvers may be necessary for the stiffest regions of large-scale
models.  Finally, we do not compare against DPM-Solver or other
structure-exploiting solvers, which occupy a different point in
the generality--efficiency trade-off.

% ============================================================
\section{Conclusion}
\label{sec:conclusion}

We derived, implemented, and benchmarked four ODE solvers for
Flow Matching.  Beyond reproducing textbook convergence rates,
we showed that the Jacobian of the learned velocity field
stiffens near $t{=}1$, that DOPRI5 adapts its step budget
accordingly, and that solver choice interacts with model maturity:
higher-order methods help most when the model is imperfect.
We hope the from-scratch implementations and the quantitative
Pareto analysis serve as a useful reference for practitioners
choosing a sampler.

% ============================================================
\bibliography{references}

% ============================================================
\newpage
\appendix

\section{Dormand--Prince Butcher Tableau}
\label{app:dopri-tableau}

\begin{equation}
\begin{array}{c|ccccccc}
0 & & & & & & & \\
\frac{1}{5} & \frac{1}{5} \\
\frac{3}{10} & \frac{3}{40} & \frac{9}{40} \\
\frac{4}{5} & \frac{44}{45} & -\frac{56}{15} & \frac{32}{9} \\
\frac{8}{9} & \frac{19372}{6561} & -\frac{25360}{2187} & \frac{64448}{6561} & -\frac{212}{729} \\
1 & \frac{9017}{3168} & -\frac{355}{33} & \frac{46732}{5247} & \frac{49}{176} & -\frac{5103}{18656} \\
1 & \frac{35}{384} & 0 & \frac{500}{1113} & \frac{125}{192} & -\frac{2187}{6784} & \frac{11}{84} \\
\hline
b_i & \frac{35}{384} & 0 & \frac{500}{1113} & \frac{125}{192} & -\frac{2187}{6784} & \frac{11}{84} & 0 \\
b_i^* & \frac{5179}{57600} & 0 & \frac{7571}{16695} & \frac{393}{640} & -\frac{92097}{339200} & \frac{187}{2100} & \frac{1}{40}
\end{array}
\end{equation}

\section{Stability Demonstration}
\label{app:stability-demo}

\begin{figure}[h]
\centering
\includegraphics[width=0.8\textwidth]{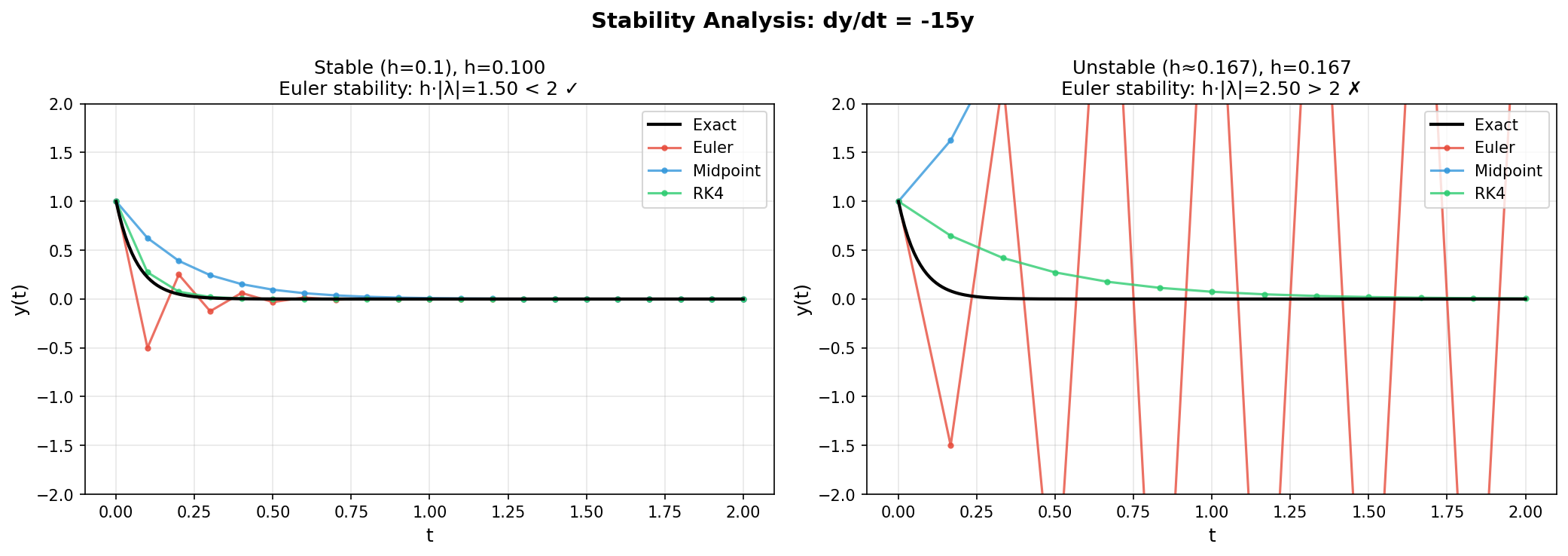}
\caption{Euler on $y'=-15y$: stable (left) vs.\ unstable (right).}
\label{fig:stability-demo-app}
\end{figure}

Figure~\ref{fig:stability-demo-app} shows Euler applied to
$y'=-15y$: stable at $h=0.1$ ($|h\lambda|=1.5<2$) and
catastrophically unstable at $h\approx0.167$ ($|h\lambda|=2.5>2$).

\section{High-Dimensional Convergence}
\label{app:highdim}

We verified identical convergence slopes in 100D and 1000D
(replicating $y'=-y$ across dimensions).  The results confirm
that our tensor-level implementation is correct; the figures
are available in the repository.

\section{Additional Datasets}
\label{app:extra}

\begin{figure}[h]
\centering
\includegraphics[width=\textwidth]{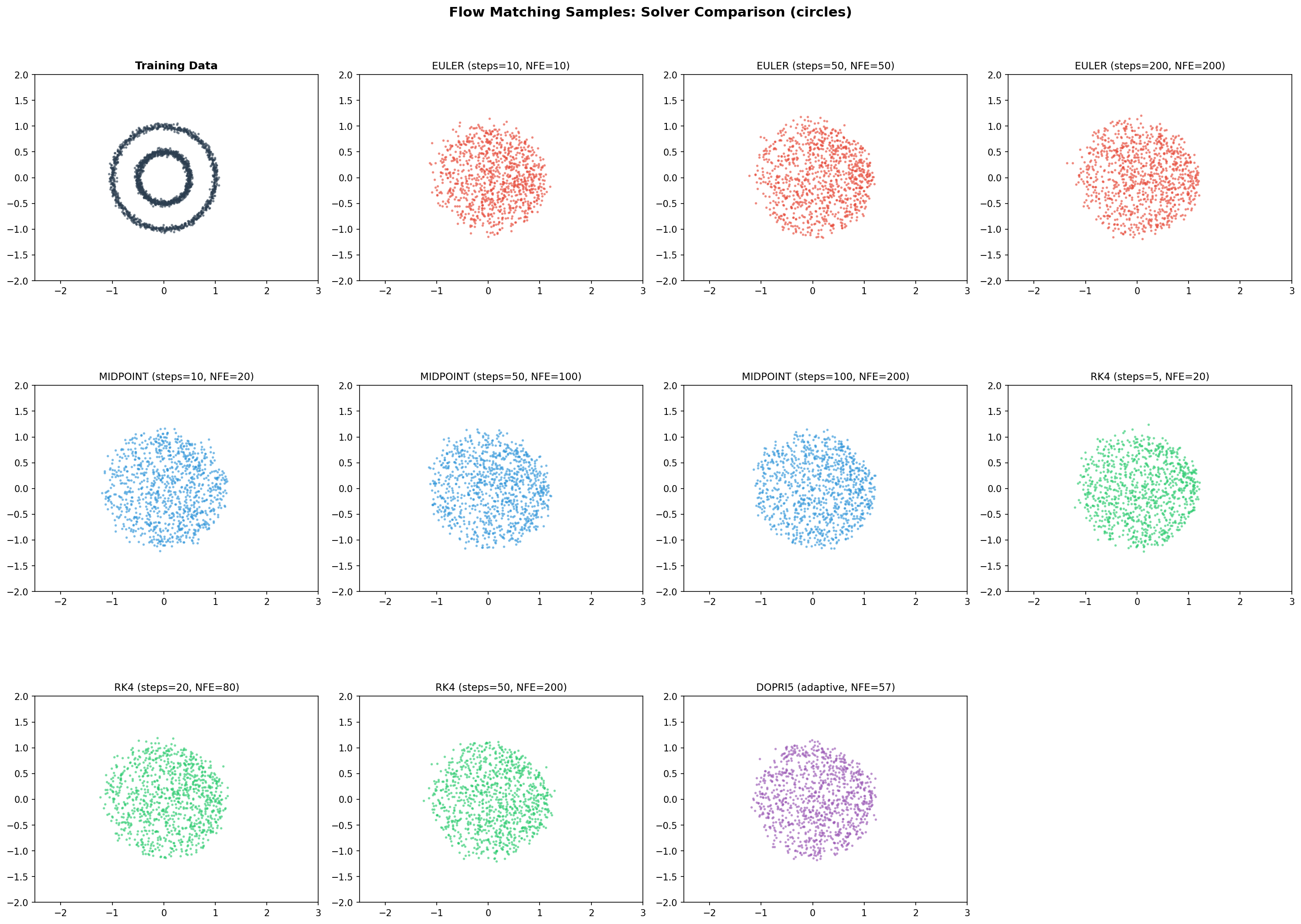}
\caption{Solver comparison on the circles dataset.}
\label{fig:circles-comparison}
\end{figure}

\begin{figure}[h]
\centering
\includegraphics[width=\textwidth]{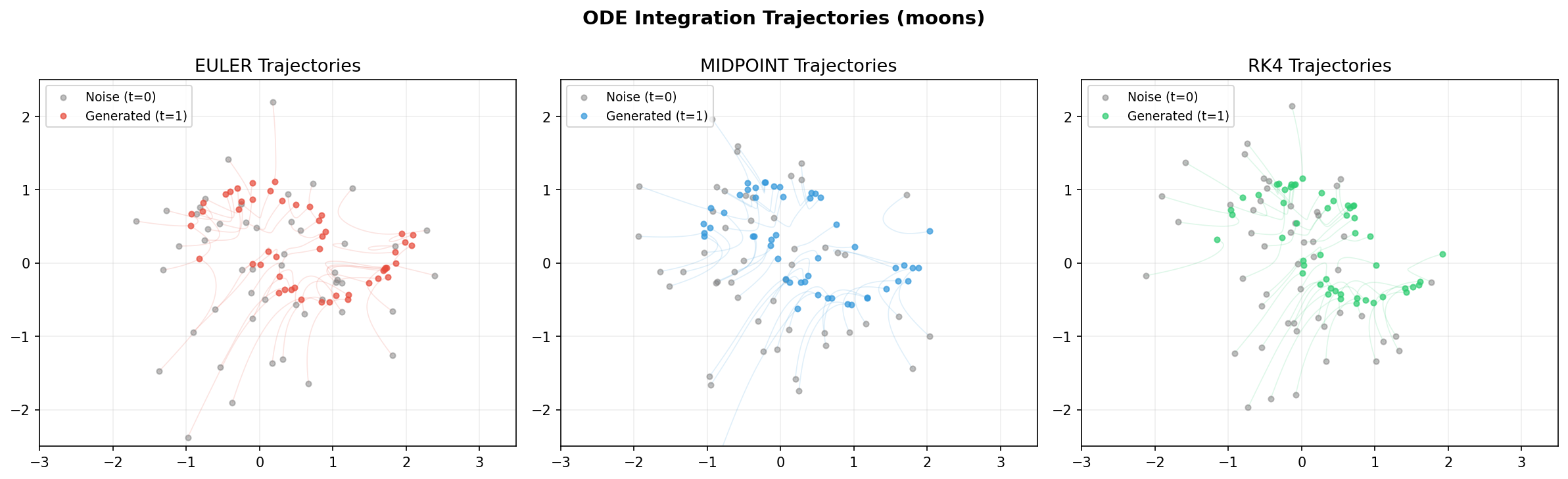}
\caption{ODE trajectories on the moons dataset (50 steps each).
Euler paths wander; RK4 paths are nearly straight---reflecting the
OT structure of the learned velocity field.}
\label{fig:trajectories-app}
\end{figure}

\section{Algorithm Pseudocode}
\label{app:algorithms}

\begin{algorithm}[h]
\caption{Euler}
\begin{algorithmic}[1]
\REQUIRE $f$, $y_0$, $[t_0,t_1]$, steps $N$
\STATE $h\gets(t_1-t_0)/N$;\; $y\gets y_0$;\; $t\gets t_0$
\FOR{$n=0$ \TO $N{-}1$}
  \STATE $y\gets y+h\cdot f(t,y)$;\; $t\gets t+h$
\ENDFOR
\RETURN $y$
\end{algorithmic}
\end{algorithm}

\begin{algorithm}[h]
\caption{RK4}
\begin{algorithmic}[1]
\REQUIRE $f$, $y_0$, $[t_0,t_1]$, steps $N$
\STATE $h\gets(t_1-t_0)/N$;\; $y\gets y_0$;\; $t\gets t_0$
\FOR{$n=0$ \TO $N{-}1$}
  \STATE $k_1\gets f(t,y)$;\;
         $k_2\gets f(t{+}h/2,\,y{+}hk_1/2)$
  \STATE $k_3\gets f(t{+}h/2,\,y{+}hk_2/2)$;\;
         $k_4\gets f(t{+}h,\,y{+}hk_3)$
  \STATE $y\gets y+(h/6)(k_1+2k_2+2k_3+k_4)$;\; $t\gets t+h$
\ENDFOR
\RETURN $y$
\end{algorithmic}
\end{algorithm}

\begin{algorithm}[h]
\caption{DOPRI5 (adaptive)}
\begin{algorithmic}[1]
\REQUIRE $f$, $y_0$, $[t_0,t_1]$, $\atol$, $\rtol$
\STATE $h\gets$ initial estimate;\; $y\gets y_0$;\; $t\gets t_0$;\;
       $k_1\gets f(t,y)$
\WHILE{$t<t_1$}
  \STATE Compute stages $k_2,\dots,k_7$ (Appendix~\ref{app:dopri-tableau})
  \STATE $y_5\gets y+h\sum b_ik_i$;\;
         $e\gets h\sum(b_i{-}b_i^*)k_i$
  \STATE $\mathrm{err}\gets\|e/(\atol+\max(|y|,|y_5|)\cdot\rtol)\|_{\mathrm{rms}}$
  \IF{$\mathrm{err}\le1$}
    \STATE $y\gets y_5$;\; $t\gets t+h$;\; $k_1\gets k_7$ (FSAL)
  \ENDIF
  \STATE $h\gets h\cdot\min(\alpha_{\max},\max(\alpha_{\min},0.9\cdot\mathrm{err}^{-1/6}))$
\ENDWHILE
\RETURN $y$
\end{algorithmic}
\end{algorithm}

\end{document}